\documentclass{llncs}
\usepackage[french,english]{babel}
\usepackage[utf8]{inputenc}
\usepackage[T1]{fontenc}
\usepackage{makeidx}
\usepackage{times}
\usepackage{amsfonts,amsmath}
\usepackage[mathcal]{euscript}
\usepackage{tikz,xcolor}
\usepackage{tkz-graph} 
\usepackage{algorithm}
\usepackage{algorithmicx}
\usepackage{algpseudocode}
\usepackage{listings}
\usepackage{caption}
\usepackage{subcaption}
\usepackage{url}
\usepackage{hhline}
\usepackage{amsmath}
\usepackage{xargs}
\usepackage{dsfont}
\usepackage{wrapfig}
\usepackage{comment}

\usepackage{multicol}
\usepackage{blindtext}

\usetikzlibrary{shapes,arrows, automata}
\usetikzlibrary{positioning}

\usepackage[colorinlistoftodos,prependcaption,textsize=tiny]{todonotes}
\newcommandx{\omer}[2][1=]{\todo[linecolor=red,backgroundcolor=red!25,bordercolor=red,#1]{#2}}
\newcommandx{\florent}[2][1=]{\todo[linecolor=blue,backgroundcolor=blue!25,bordercolor=blue,#1]{#2}}
\newcommandx{\tianqi}[2][1=]{\todo[linecolor=green,backgroundcolor=green!25,bordercolor=green,#1]{#2}}
\newcommandx{\stefan}[2][1=]{\todo[linecolor=orange,backgroundcolor=orange!25,bordercolor=orange,#1]{#2}}

\title{ICTAC2020}
\author{Omer Nguena Timo}
\date{June 2020}

\begin{document}

\title{An Approach to Evaluating Learning Algorithms for Decision Trees}

\author{Tianqi Xiao\inst{1} \and Omer Nguena Timo\inst{2} \and Florent Avellaneda\inst{2} \and Yasir Malik\inst{3} \and Stefan Bruda\inst{1}}

\institute{
Bishop's University, Sherbrooke (QC), Canada\\
\email{xiao@cs.ubishops.ca, stefan@bruda.ca}
\and 
CRIM - Computer Research Institute of Montreal,
Montreal (QC), Canada \\   \email{\{omer.nguena-timo, florent.avellaneda\}@crim.ca}
\and
New York Institute of Technology, Vancouver (BC), Canada\\
\email{ymalik@nyit.edu}
}
\maketitle

\begin{abstract}
Learning algorithms produce software models for realising critical classification tasks. Decision trees models are  simpler than other models such as neural network and they are used in various critical domains such as the medical and the aeronautics. Low or unknown learning ability  algorithms does not permit us to trust the  produced software models, which lead to costly test activities for  validating the models and to the waste of learning time in case the models are likely to be faulty due to the learning inability.  
%They are easy to read and understand, which renders them  more popular than  other models such neural networks. 
%
Methods for evaluating the decision trees learning ability, as well as that for the other models, are needed especially since the testing of the learned models is still a hot topic.  
%
%One reason is the difficulty to estimate the quantity and quality of available data needed for the evaluation learning and testing of the trees. Therefore, the results obtained by classical techniques to evaluate learning algorithms, such as cross-validation, will depend on the quality of these data and not only on the quality of the learning algorithm that we want to evaluate.
%Moreover of traditional metric-based evaluation techniques apply on specific learned trees; they cannot be used to assess the correctness of the learning algorithms which produce the learned trees.
We propose a novel oracle-centered approach to evaluate (the learning ability of) learning  algorithms for decision trees. It consists of generating data from reference trees playing the role of oracles, producing learned trees with existing learning algorithms, and determining the degree of correctness (DOE) of the learned trees by comparing them with the oracles. The average DOE is used to estimate the quality of the learning algorithm.  the We assess five decision tree learning algorithms based on the proposed approach.   

\end{abstract}

\section{Introduction}
Binary decision trees are algorithms used to assign outputs (classes) to inputs (values of feature variables). This classification task is performed via  decisions on the inputs; it  can be critical in some application domains such as the medical (e.g., diagnosing breast cancer \cite{rajesh2012analysis}) 
%or COVID-19 \cite{shi2020large} from patients' medical parameters/data)
, home automation~\cite{AshokJJKWZ20a} and  the aeronautics domain (e.g., airborne collision avoidance systems \cite{PedregosaVGMTGBPWDVPCBPD11}). Decision trees are easy to read, understand, and interpret; they can be preferred to  other machine learning models such as neural networks. Decision trees can be automatically designed using learning algorithms. Learning algorithms take as an input a training dataset consisting of input-output pairs and produce learned decision trees; it is desired that the learned trees generalize the given training dataset, i.e., compute or predict the expected outputs for new inputs. In this paper, we  go beyond the evaluation of the quality of decision trees produced by applying learning algorithms on domain-specific datasets and our goal is evaluating the quality of the learning algorithm themselves.

Decision tree learning algorithms iterate on the "best" splitting of the training dataset until a stopping criterion is reached. So, defining  new splitting and stopping criteria give rise to new learning algorithms.  Cross-validation has been adopted to evaluate the adequacy of learning algorithms to produce  "accurate" decision trees for classification problems. Accuracy is estimated with metrics such as precision, recall, F1 score~\cite{Chinchor92}. For example, the precision is the ratio of test data that are correctly classified by the learned tree. Both the cross-validation and accuracy depend on existing  training and test datasets. We believe that this is a weakness of the existing evaluation methods because the results obtained with a specific dataset are generalized to the input space of the decision trees. It is not obvious that if a learning algorithm produces the best cross-validation score with a specific dataset, it means that the algorithm is best for learning decision trees. Indeed, the cross- validation technique use parts of the given specific dataset as validation datasets and it is not immediate that the given dataset is representative enough of the unknown artefact which has produced the dataset.  Nothing ensures that the cross-validation score will remain the best in case  the dataset is changed.  The  classical evaluation techniques for the quality of  decision tree learning algorithms, such as cross-validation,  depend on the quality of chosen datasets. We believe that this is a limitation that we overcome by  proposing a statistic-based quality evaluation approach. The unknown artefact is in fact an oracle for the learned decision trees and it can be used to evaluate the learned trees; then a sufficiently large amount of oracles can serve to evaluate whether or not a learning algorithm realises the expected learning task, i.e., it is "correct". 

Our contribution is an oracle-centered approach for a statistical evaluation of the  quality of decision tree learning algorithms and an emprirical assessment of the approach with five existing decision tree learning algorithms. The approach does not depend on any domain-specific dataset and it is based on a comparison between  randomly generated decision tree oracles and learned trees, which is a novelty. Randomly generated oracles serve to generate only training datasets are given as inputs to learning algorithms. We consider two dataset generation modes. We introduce a new metric called the the degree of equivalence (DOE) between learned trees and oracles; it serves to evaluate the quality of the learned trees and the learning algorithms.  Within the evaluation approach, we specifically establish:
\begin{itemize}
    \item relations between the size of the training datasets and the degree of equivalence of the learned decision trees,
    \item an evaluation of the quality of decision tree learning algorithms in terms of the best (average) degree of equivalence,
    \item a comparison of the evaluation results for the two dataset generation modes. 
\end{itemize}

The rest of the paper is structured as follows. In Section~\ref{sec:prelim}, we summarize and present the necessary preliminaries, including the definition of decision tree and its properties. We also discussed the existing evaluation metrics which are commonly used in analyzing the degree of correctness of the decision trees. In Section~\ref{sec:dist-tree}, we describe the equivalence relation between decision trees by introducing distinguishing trees. In Section~\ref{sec:approach} we propose  a novel evaluation approach over distinguishing trees and explain the design of the approach in detail. We then assess  our approach with empirical experiments and discuss the results in Section~\ref{sec:exp}. We finally conclude the study in Section~\ref{sec:concl}.

\section{Preliminaries}\label{sec:prelim}

\subsection{Decision Trees}
A decision tree is an algorithm for making decisions based on the values of some input parameters called \textit{features}. It can be represented with a tree structure also called a decision tree. We define it as a graph that satisfies some properties. 

\par A \textit{(labeled) graph} (shortly, a graph) is a tuple $T = (N, n^0, E, F, V, lab_N)$ where $N$ is a finite set of nodes, $n^0\in N$ is the initial node, $V$ is a finite set of labels for the edges, $E\subseteq N \times V \times N$ is a finite set of labeled edges, $F$ is a finite set of labels for the nodes,  $lab_N: N \to F$ is a labeling function for the nodes. Given an edge $e=(n,v, n')$, $src(e)=n$, $lab_E(e) = v$ and $tgt(e)=n'$ are the source, the label and the target of $e$; $e$ is called an entering (resp. leaving) edge for $n'$ (resp. $n$). A path of length $l$ from node $n$ in  $T$ is a sequence of edges $\pi = e_1\ldots e_l$ such that $n=src(e_1)$,$tgt(e_i) = src(e_{i-1})$ for every $i=1..l-1$; $\pi$ is acyclic if $tgt(e_j)\neq src(e_k)$ for every $j,k$ such that $j \geq k$. We say that $T$ is a (labeled) \textit{tree} if it satisfies the following properties: every path in it is acyclic, each node except the root has exactly one entering edge and there is a unique path from the root $n^0$ to each node. Given an edge $e=(n, v, n')$ of a tree $T$, $n'$ (resp. $n$) is called the \textit{child} (resp. \textit{parent}) of $n$ (resp. $n'$) and $v$ is the \textit{label} of $e$. We also define the descendant and the ancestor relation as the transitive closure of the child and parent relation. A node with no descendant is called a \textit{leaf}; all the other node are called \textit{internal}.   The depth of a node $n$ in a tree $T$, denoted by $depth(n)$, corresponds to the length of the only path from the root to the node $n$. The depth of $T$, $depth(T)$ is the maximal depth of its leaves. A \textit{binary tree} is a tree such that each internal node has two children called the left child and the right child.

\par A \textit{decision tree} is a tree $T = (N, n^0, E, F \cup C, V, lab_N)$ where $F=\{f_1,f_2,\ldots,f_n\}$ represents a \textit{feature} set, $C=\{c_1,c_2,\ldots c_k\}$ represents a \textit{class} set and $V = V_{f_1}\cup V_{f_2}\cup \ldots\cup V_{f_n}\}$ represents feature value sets, $f_i$ takes its values in  $V_{f_i}$, $i=1...n$. $T$ also satisfies the following conditions: $lab_E(e) \neq lab_E(e')$ for every pair of edges $e$ and $e'$ leaving the same node; $lab_N(n)\in C$ if $n$ is a leaf; two nodes in a same path from the root to a leaf have distinct labels; 
and each label in $V_i$ is assigned to an edge leaving  from each internal node labeled with $f_i$. If the size of $C$ equals $2$, i.e., $|C|=2$, then $T$ is called a binary decision tree. 
Intuitively, the label of an internal node represents a feature, the label of a leaf represents  a \textit{class}, the label $v$ in an edge $(n,v,n')$ represents a value of the feature $lab_N(n)$. We let $rule(T)$ denote the set of paths from root $n^0$ to a leaf and each element in $rule(T)$ is called a \textit{rule}.
We let $feature(\pi)$  represents the set of features labeling the nodes of rule $\pi$. Note that $feature(\pi)\subseteq F$, meaning that the values of some features can be left undefined in $\pi$. The input for $\pi$ corresponds to the values assigned to the feature in $feature(\pi)$ along $\pi$.
A rule represents the class that holds on the input. A feature with an undefined value in $\pi$ can take any value without changing the class. 
$T$ is called a \textit{decision binary-tree} if $|V_f|=2$ for every $f\in F$. Two decision trees are \textit{equivalent} if they make an identical decision for each evaluation of the features. Note that each decision tree can be represented with an equivalent decision binary-tree. Henceforth, every decision tree is a decision binary-tree.

\par
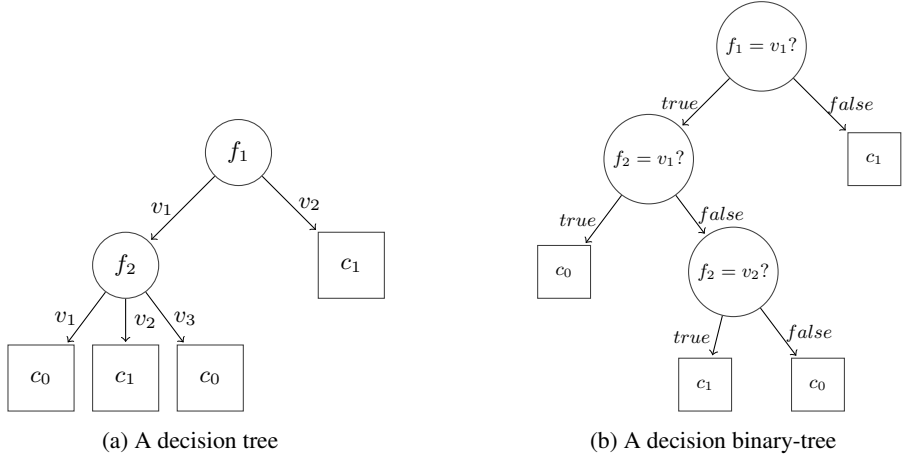
\begin{figure}[t]
 \centering
  \subfloat[A decision tree]{
     \label{fig:ex1:dt}
     \scalebox{1}[1]{
     \begin{tikzpicture}[scale=1.5,shorten >=1pt,->,every node/.style={scale=1}]
    \tikzstyle{vertex}=[circle,draw=black!75,minimum size=25pt]
    \tikzstyle{leaf} = [rectangle,draw=black!75,minimum size=25pt]
     \node[vertex] (Q1) at (0,0) 	{$f_1$}  ;
     \node[vertex] (Q2) at (-1,-1) 	{$f_2$}  ;
     \node[leaf] (Q3) at (1,-1) 	{$c_1$}  ;
     \node[leaf] (Q4) at (-1.75,-2) {$c_0$}  ;
     \node[leaf] (Q5) at (-1,-2)	{$c_1$}  ;
     \node[leaf] (Q6) at (-0.25,-2) 	{$c_0$}  ;
     
     \path (Q1) edge node[left]         {$v_1$}  	(Q2);
     \path (Q1) edge node[right]         {$v_2$}  	(Q3);
     \path (Q2) edge node[left]         {$v_1$}  	(Q4);
     \path (Q2) edge node[right]         {$v_2$}  	(Q5);
     \path (Q2) edge node[right]         {$v_3$}  	(Q6);
     \end{tikzpicture}
     }}
     \hfill
  \subfloat[A decision binary-tree]{
    \label{fig:ex1:dbt}
    \scalebox{1}[1]{
    \begin{tikzpicture}[scale=1.5,shorten >=1pt,->,every node/.style={scale=0.8}]
    \tikzstyle{vertex}=[circle,draw=black!75,minimum size=25pt]
    \tikzstyle{leaf} = [rectangle,draw=black!75,minimum size=25pt]
 
    \node[vertex] (Q1) at (0,0) 	{$f_1 = v_1?$}  ;
    \node[vertex] (Q2) at (-1,-1) 	{$f_2 = v_1?$}  ;
    \node[leaf] (Q3) at (1,-1) 	{$c_1$}  ;
    \node[leaf] (Q4) at (-1.75,-2) {$c_0$}  ;
    \node[vertex] (Q5) at (-0.25,-2)	{$f_2 = v_2?$}  ;
    \node[leaf] (Q6) at (-0.5,-3) 	{$c_1$}  ;
    \node[leaf] (Q7) at (0.5,-3) 	{$c_0$}  ;
    
    \path (Q1) edge node[left]         {$true$}  	(Q2);
    \path (Q1) edge node[right]         {$false$}  	(Q3);
    \path (Q2) edge node[left]         {$true$}  	(Q4);
    \path (Q2) edge node[right]         {$false$}  	(Q5);
    \path (Q5) edge node[left]         {$true$}  	(Q6);
    \path (Q5) edge node[right]         {$false$}  	(Q7);
     
    \end{tikzpicture}
    }}
  \caption{Examples of decision trees}
  \label{fig:ex1}
\end{figure}

 Figure~\ref{fig:ex1} presents two equivalent decision trees. The decision tree in Figure~\ref{fig:ex1:dbt} is binary whereas the one in Figure~\ref{fig:ex1:dt} is not. This latter tree  has $5$ nodes including two internal nodes and four leaves. The names of the  nodes are missing in the figure; only their labels are shown. The label of the root is feature $f_1$. The decision tree in Figure~\ref{fig:ex1:dt} allows classifying feature vectors into two classes, namely $c_0$ and $c_1$. The leftmost rule of the tree indicates the class $c_0$ corresponds to the assignment of $v_1$ to $f_1$ and $f_2$. The value of $f_2$ is not defined in the rightmost rule, i.e., $c_1$ is decided class when $f_1=v_2$ whatever is the value of $f_2$.

\subsection{Decision Trees Consistent with Datasets}
A  dataset is nothing else but a set of input-output pairs. Inputs are values of features and outputs are classes for the inputs.  
Formally, let $F$ be the set of features, $Dom_F$ be the domain for $F$. In this work $V_F= \{true,false\}$ since we consider binary-tree as discussed in Section~\ref{sec:prelim}. An \textit{input} $x$ is a valuation of $F$, i.e., a total function $x: F \to V_F$. Then, $x[f:=v]$ is the valuation obtained from $x$ by assigning $v$ to $f$ and letting the value of the other feature unchanged. We let $(V_F)^F$ denote the set of all possible inputs. The tuple $(x,c)$ is called a \textit{dataset instance}. Let $C$ be a set of possible  classes for the inputs. In this work we consider deterministic datasets, i.e., at most one output corresponds to each input. A (deterministic) dataset is a partial function $D : (V_F)^F \to C \times \mathbb{N}$ which assigns a class to inputs and a redundancy factor to the assignment; in other words, $D(x) = (c,k)$ indicates that the label $c$ is assigned to $x$ and this assignment occurs $k$ times in $D$. Note that since $D$ is a partial function, not every input is assigned a class. We let $V_D$ represent the set of inputs for which $D$ assigns a class. The size of $D$ is the sum of the redundancy factors for the inputs on which $D$ is defined. 

 Let $D$ be a dataset, $x$ be an input such that $D(x) = (c,k)$ and $\pi$ be a rule of the decision tree $T$. We say that $x$ \textit{triggers} $\pi$ if the projection of $x$ on $feature(\pi)$ corresponds to the input of $\pi$, i.e.,  $x$ and the input of $\pi$ assign the same value to each feature $feature(\pi)$; moreover if the class for $\pi$ equals $c$ then we say that instance $(x,c)$ is consistent with $\pi$   and we write $\pi(x) = c$. Instance $(x,c)$ is  \textit{consistent with}  $T$  if  it is consistent with some rule  in $T$ and we write $T(x) = c$. Dataset $D$  is \textit{consistent} with  $T$  if all its instances are consistent with $T$. The \textit{precision} of $T$ on $D$ is the ratio of the number of inputs in $D$ with which $T $ is consistent over the size of $D$.
 Note that there is at most one consistent rule for a given instance (input). A rule $\pi$ can be consistent with many distinct instances; this is because $\pi$ does not define the features in  $F\setminus feature(\pi)$ and undefined features can take any values. 
 
 Consider the decision tree in Figure~\ref{fig:ex1:dt} and the datasets $D_1$ and $D_2$ presented in Table~\ref{tab:ex:dataset:d1} and Table~\ref{tab:ex:dataset:d2} 
 
 \begin{table}[!t]
  \centering
  \subfloat[Dataset $D_1$, where $D_1(v_2,v_1) = (c_1,1), D_1(v_1,v_3) = (c_0,2)$]{
  \begin{minipage}[c][0.5\width]{
	   0.45\textwidth}
   \label{tab:ex:dataset:d1}
  \centering
     \begin{tabular}{|c|c||c|}
        $f_1$   &  $f_2$ & class \\ \hline
        $v_2$  & $v_1$ & $c_1$\\
        $v_1$ & $v_3$ & $c_0$ \\
        $v_1$ & $v_3$ & $c_0$ \\
     \end{tabular}
     \end{minipage}
     }%
     \hfill
  \subfloat[Dataset $D_2$]{
   \begin{minipage}[c][0.5\width]{
	   0.45\textwidth}
  \label{tab:ex:dataset:d2}
  \centering
     \begin{tabular}{|c|c||c|}
        $f_1$   &  $f_2$ & class \\ \hline
        $v_2$  & $v_1$ & $c_1$\\
         $v_2$  & $v_1$ & $c_1$\\
          $v_2$  & $v_1$ & $c_1$\\
        $v_1$ & $v_3$ & $c_1$ \\
        $v_1$ & $v_3$ & $c_1$ \\
     \end{tabular}
     \end{minipage}
 }
  \caption{Examples of datasets}%
  \label{tab:ex:dataset}%
\end{table}

% represented with tuples of valuations of the features and the corresponding classes: $D_1 =\{(f_1=v_2, f_2=v_1, (c_1,1)), (f_1=v_1, f_2=v_3, (c_0,2))\}$ and $D_2 =\{(f_1=v_2, f_2=v_1, (c_1,3)), (f_1=v_1, f_2=v_3, (c_1,2))\}$. 
Dataset $D_1$ is consistent with the decision tree and $D_2$ it is not consistent with the decision tree  because   $D_2(x) = (c_1,2)$ with $x=(f_1=v_1, f_2=v_3)$ and the decision $c_0$ for the rule $f_1\xrightarrow{v_1}f_2\xrightarrow{v_3}c_0$ triggered by $x$ differs from the class $c_1$ for $x$ in $D_2$. The precision of the tree on $D_1$ and $D_2$ are $100\%$ and $50\%$, respectively. 
 
\subsection{Learning and Evaluation of Decision Trees}
A learning algorithm $A$ for decision trees take as an input a so-called learning dataset $D_l$ and produces a learned decision tree $T$. Most of the Decision tree learning algorithms work in three main steps~\cite{breslow1997simplifying,Breiman1984,quinlan1986induction,Quinlan1993,frank2009weka,DBLP:conf/aaai/Avellaneda20}.
The first step corresponds to iterative splitting of instances of a current dataset into subsets.  The splitting is done according to a criterion on a particular feature. The initial dataset is $D_l$ and each new subset becomes the initial dataset for a next iteration. The splitting and the iteration stopping criteria characterize the algorithms. A preliminary decision tree is produced at  the end of the splitting step. The second step corresponds to the pruning of the preliminary decision; it consists in assigning classes to internal nodes of the preliminary tree and to remove their descendants. The result of this step a new tree having a smaller size than the preliminary tree. This step is motivated by the parsimony principle stating that "small models that are consistent with the observations are more likely to be the right one". In our context, smaller (simpler) trees could generalize better than the bigger (complex) ones.
In case there are several smaller tree candidates, one of them can be selected by applying adequate techniques, e.g.,the cross-validation, in the last step of the learning algorithm.

We want to evaluate the degree of correctness of  decision trees and increase our confidence in them prior to  rolling them out. First, we must choose the "best" learning algorithm and secondly, we should assess the performance of learned trees produced by the chosen algorithm. The most popular technique for choosing the "best" learning algorithm is cross-validation~\cite{refaeilzadeh2009cross}. Given a collection of $l$ learning algorithms $\{A_i\}_{i=1..l}$ and a training dataset $D_t$ representing a classification problem, the cross-validation technique works as follows. First $D_t$ is partitioned into $k$ subset $\{D_{t_i}\})i=1..k$. Each algorithm $A_i$ is executed $k$-times by considering the union of $k-1$ subsets as the training dataset and the remaining subset as a test dataset. Each application produces a learned decision tree of which the precision is determined with the test dataset. The global precision for $A_i$ is the average of the $k$ precision measures obtained with the $k$ test subsets. In the end, the "best" algorithm is the one with the higher global precision and it is chosen as well as one of the learned trees it has produced. A similar statement holds for other metrics (e.g.,  recall, F1 score, confusion matrices~\cite{Chinchor92})  used  to evaluate the quality of learned trees. Note that a chosen learning for a classification task is adopted by  learning communities and applied on different  datasets, e.g., provided by different actors wanting to perform the same classification but from different or private  databases. The results of the cross-validation depend on the quality of the datasets and not only on the quality of the learning algorithms; this is a limitation. \\ 

Our goal is proposing an evaluation technique for decision tree learning algorithms that is independent of a specific dataset. Our technique is statistic-based; it uses (randomly generated) reference trees as oracles, learns new decision trees from datasets produced with oracles, and estimates the difference between learned trees and oracles.  

\section{Estimating the Difference Between Two Decision Trees}
\label{sec:dist-tree}

We introduce an equivalence relation between decision trees and a notion of distinguishing tree to check the equivalence of two decision trees and to estimate the difference between decision trees.  Let $T_1 = (1  , n_1^0, E_1, F_1\cup C_1 , V_1 , lab_{N_1})$ and $T_2 = (N_2  , n_2^0, E_2, F_2 \cup C_2 , V_2 , lab_{N_2})$ be two decision trees. Without loss of generality, we assume that $F_1= F_2 =F$ and $C_1= C_2 =C$. We recall that $T_1(x)$ represents the label/class the tree $T_1$ produces for input $x$.

\subsection{Decision Tree Equivalence and Distinguishing Tree}

    \begin{definition}\label{def:equivalent}
A tree $T_1$ is \emph{equivalent} to a tree $T_2$ if for each input $x: F\to V_{F}$ we have $T_1(x) = T_2(x)$; otherwise there is an input $x$ such that  $T_1(x) \neq T_2(x)$ and $x$ is called a \textit{distinguishing input}.
    \end{definition}

\par We would like to estimate the difference between $T_1$ and $T_2$ in case they are not equivalent. A naive way to check the equivalence is to enumerate all possible inputs and check for how many $T_1(x) = T_2(x)$. However, this method is not efficient because the number of distinct inputs is exponential in the number of features. We will rather  compare their rules and ensure that if the input $x$  of a rule $\pi_1$ of $T_1$ (resp. $\pi_2$  of $T_2$) triggers a rule $\pi_2$  of $T_2$ (resp. $\pi_1$  of $T_1$), then both rules determine the same class, i.e., $\pi_1(x) = \pi_2(x)$. Because the number of rules is fewer than the number of instances (recall that some features might not be defined in rules), the rule comparison approach can be more efficient in practice. 

\par To perform the rule comparison, we define a notion of the distinguishing tree in a constructive manner. It can be seen as a product (parallel execution) of two decision trees and  is inspired by the definitions in ~\cite{Nguena-TimoPR18,TimoPR19,timo2019using} of distinguishing automata for  extended finite state machines. Note that features do not necessarily occur at the same depth in the two decision  trees; this justifies the introduction of inputs (used to keep track of  the values of encountered features) in nodes of distinguishing trees, in addition to the nodes of the two decision trees.

\begin{definition} 
    Given two decision trees $T_i =  (N_i  , n_i^0, E_i, F_i \cup C_i , V_i , lab_{N_i})$ with $i=1,2$ and such that $F_1 =F_2$, $C_1=C_2$ a tree $T = (N\cup \nabla, n^0, E, F \cup C , V , lab_{N})$ where $V= V_1\cup V_2$, $F= F_1 = F_2$, $C= C_1 = C_2$,  $N\subseteq N_1 \times N_2 \times (V_{F})^F$, $E \subseteq N \times V\times N$, $\nabla \subseteq N$ is the subset of accepting nodes, is a \textrm{distinguishing tree} for $T_1$ and $T_2$ if it holds that :
    \begin{itemize}
        \setlength\itemsep{0.5em}
        \item[$(\mathcal{R}_0)$] $n^0 = (n_1^0, n_2^0, x_\bot)$ is the root with $x_\bot$ a partial input that assigns an undefined value to each feature, i.e., $x(f) = \bot$ for every $f\in F$
        
        \item[$(\mathcal{R}_1)$] $((n_1, n_2, x), v, (n'_1, n'_2, x[f := v])) \in E$ if $f = f_1 = f_2 = lab_{N_1}(n_1) = lab_{N_2}(n_2)$, $(n_1, v_1, n'_1) \in E_1$, $(n_2, v_2, n_2') \in E_2$ and $v=v_1=v_2$.
    
        \item[$(\mathcal{R}_2)$] $((n_1, n_2, x), v_1, (n'_1, n_2, x[f_1 := v_1])) \in E$ if $f_1 \neq f_2$ with $f_1 = lab_{N_1}(n_1)$, $f_2 = lab_{N_2}(n_2)$ and $(n_1, v_1, n'_1) \in E_1$.
    
        \item[$(\mathcal{R}_3)$] $((n_1, n_2, x), v_2, (n_1, n'_2, x[f_2 := v_2])) \in E$ if $f_1 \neq f_2$ with $f_1 = lab_{N_1}(n_1)$, $f_2 = lab_{N_2}(n_2)$ and $(n_2, v_2, n'_2) \in E_2$.
        
        \item[$(\mathcal{R}_4)$]  $(n_1, n_2, x)\in \nabla$ iff $n_1$ and $n_2$ are leaves and correspond to distinct  classes, i.e., $lab_{N_1}(n_1)\neq lab_{N_2}(n_2)$. The leaves in $\nabla$ are called accepting.
    \end{itemize}
\end{definition}

\par The root node of $T$ corresponds to the case where  $T_1$ and $T_2$ are positioned at their roots ($(\mathcal{R}_0)$) and $x_\bot$ indicates that no feature is assigned to a value.
From a node $n$ of $T$, there are three cases which correspond to rules $(\mathcal{R}_1)$, $(\mathcal{R}_2)$ and $(\mathcal{R}_3)$. In $(\mathcal{R}_1)$, the nodes $n_1$ and $n_2$ correspond to the same feature $f$ and were assigned to the same value $v$; in this situation, we move to the children of the nodes and update the input $x$ by setting the value of $f$  to $v$. In $(\mathcal{R}_1)$ and In $(\mathcal{R}_2)$, the features for $n_1$ and $n_2$ are different; then we can move to the child of $n_1$  while staying in $n_2$ (see  $(\mathcal{R}_1)$) and to the child of $n_2$ while staying in $n_1$ (see  $(\mathcal{R}_2)$); the input $x$ is updated  in either case. Accepting nodes in $\nabla$ contain  leaves of $T_1$ and $T_2$ with different classes (see $(\mathcal{R}_4)$).  

Note that if $(n_1,n_2,x)$ is accepting, i.e., it belongs to $\nabla$, then $x$  is a distinguishing input for $T_1$ and $T_2$.

A path in $T$ from the root to an accepting leaf is called accepted. We let $Acc(T)$ denote the set of accepted path of $T$.
   \begin{proposition}\label{prop:distinguishing tree} 
        Let $T$ be a \emph{distinguishing tree} for $T_1$ and $T_2$. It holds that:
        \begin{enumerate}
            \item For each  path $\pi \in Acc(T)$, if an input $x$ triggers $\pi$, then $T_1(x) \neq T_2(x)$.
            \item Each input $x$ such that $T_1(x) \neq T_2(x)$ triggers a path $\pi \in Acc(T)$.
        \end{enumerate}
    \end{proposition}

 \begin{proposition}
        $T_1$ and $T_2$ are  equivalent if and only if $Acc(T) =\emptyset$, where $T$ is the distinguishing tree for decision trees $T_1$ and  $T_2$.
\end{proposition}

\subsection{Degree of Equivalence}
We introduce the degree of equivalence (DOE) between two decision trees, which we use to estimate the difference between them. The DOE is defined based on the consistency between rules rather than the consistency between instances. A major difference between rules and instances is that rules do not necessarily define all the features whereas instances do. As a consequence, the DOE based on rule counting differs from the precision which is based on instance counting. Defining DOE is relevant in our work because we use the oracle tree as a reference for evaluating learned trees and both trees provide rules.

\par Let $\pi_1$ be a rule in $T_1$  with input $x_{\pi_1}$ and $\pi_2$ be a rule in $T_2$ with input $x_{\pi_2}$. We say that $\pi_1$ is \textit{compatible} with $\pi_2$ if some input $x \in V_F$ triggers both $\pi_1$ and $\pi_2$. 
We say that $\pi_1$ is \textit{consistent} with $\pi_2$ if both rules are compatible and  $\pi_1(x) = \pi_2(x)$, i.e., the two rules give the same class for $x$. If an input $x$ triggers two consistent rules  $\pi_1$ and $\pi_2$, then the projection of $x$ on $feature(\pi_1)$ (resp. $feature(\pi_2)$) equals $x_{\pi_1}$ (resp. $x_{\pi_2}$). 

\begin{definition}   The \emph{degree of equivalence of a tree $T_1$ w.r.t. a decision tree $T_2$} is the  ratio of the rules of $T_1$ consistent with a rule $\pi_2$ of $T_2$ over the  total number of rules of $T_1$ compatible with a rule $\pi_2$ of $T_2$.
\end{definition}

The DOE of $T_1$ w.r.t $T_2$ is a rational number between $0$ and $1$; this is simply because consistent rules are included in compatible rules. If $T_1$ and $T_2$ are equivalent then the DOE of $T_1$ w.r.t. $T_2$ equals $1$. 

\par Every  path  $\pi$ to a leaf in the distinguishing tree $T$ for $T_1$ and $T_2$ determines a tuple of compatible rules $(\pi_1,\pi_2)$ such that $\pi_1$ is a rule of $T_1$ and $\pi_2$ is a rule of $T_2$. Then the two rules are consistent if and only if  $\pi$ is not accepting. This suggests that the distinguishing tree can be used to determine the DOE. Later in Section~\ref{sec:approach:doe}, we devise an algorithm for building the distinguishing tree meanwhile computing the DOE.

\section{An Approach to Evaluate Decision Tree Learning Algorithm}
\label{sec:approach}
Based on the idea of distinguishing trees and the DOE metric, we proposed an approach to evaluating the performance of decision tree learning algorithms empirically. This approach consists of three components: an oracle generator that randomly builds oracle trees, a training dataset generator that randomly produces datasets from the generated oracle trees, and an equivalence tester that compares the trained model with the oracle and calculates the DOE value for the trained trees w.r.t. oracles. 

\subsection {Decision Tree Oracle Generation}
To produce the oracle tree, the generator takes as inputs some parameters from the user. The parameters are the desired number of features $m$, the depth of the rules $k$, and the size  $Z_i$ for the domain $V_{f_i}$ of each feature $f_i$, $i=1..m$. As we explained in Section~\ref{sec:prelim}, we consider decision binary-tree, without the loss of generality. To better demonstrate the usage of the generator, we design the generator to always produce "perfect" trees,  which means that all the rules described by the tree will have an identical number of features. 
Note that though the length (depth) of each rule is the same, if the depth $k$ is smaller than $m$, each rule may use different feature subsets. In Figure~\ref{fig:perfect tree}, though the depth of the leftmost rule and the rightmost rule is the same ($3$), the feature subset of the leftmost rule ($f_2, f_3$) is different from the feature subset of the rightmost rule ($f_3, f_4$). 

\begin{algorithm}[tb]{\textbf{Input:} $m$, $k$, $\{Z_i\}_{i=1..m}$
\\
\textbf{Output:} $O$, an oracle tree}
\begin{multicols}{2}
    \setlength\columnsep{30pt}
    \begin{algorithmic}[1]
        \State Create $F$ with $m$ features $\{f_0, f_1, ..., f_{m-1}\}$
        \State Create $C$ the set of classes
        \For{each $f_i$ in $F$}
            \State Create $V_{f_i}$ with $Z_i$ values
        \EndFor
        \State Create root node $n_0$ in $O$
        \If{$|F| \neq 0$}
           % \State $lvl \leftarrow 0$ \Comment{$lvl$ denotes depth of $O$}
           % \State \Call{expandtree}{$t_0$, $F$, $lvl$}
           \State \Call{expandtree}{$n_0$, $F$}
        \EndIf
        \State \textbf{return} $O$ 
   \\
    \Procedure{expandtree}{$n$, $F_s$} \Comment{$F_s \subseteq F$}
        \If{$|F_s| \neq 0$ \textbf{and} depth($n$) $<$ $k$}
            \State $i \leftarrow$ random ($0$, $m-1$)  \Comment{choose random from $F_s$}
            \State $lab_N(n) \leftarrow f_i$
            \For{each $v$ in $V_{f_i}$}
                \State Create a new node $n_{new}$
                \State create an edge $(n, v, n_{new})$ in $O$ \Comment{add  $n_{new}$ as a child of  $n$}
                %\State \Call{expandtree}{$t_l$, $F_s \setminus f_i$, $lvl+1$} \Comment{"$\setminus$" denotes exclusion}
                 \State \Call{expandtree}{$n_{new}$, $F_s \setminus f_i$} \Comment{"$\setminus$" denotes exclusion}
            \EndFor
         %\ElsIf{depth($O$) $=$ $k$}
            %\For{each $j$ in $|C|$}
               % \State create $t_j$ $\rightarrow$ children($t$)
           % \EndFor
        \ElsIf{depth($n$) $=$ $k$}
        \State $c\leftarrow$  clever\_random\_from($C$)
        \State $lab_N(n) \leftarrow c$
        \EndIf
    \EndProcedure
    \end{algorithmic}
    \caption{Oracle Tree Generator}
    \label{algo: rule generator}
    \end{multicols}
\end{algorithm}

Algorithm~\ref{algo: rule generator} presents the pseudo-code of the generator. The generator starts by initializing the feature set $F$ and assigns values to each feature.
Then, the generator creates the root node of the oracle and randomly assigns a feature to it. The tree expands from the root node by adding nodes recursively and choosing features in a random fashion until the prescribed depth $k$ is reached. Each node is labeled with one feature which has not yet been selected by its ancestors, because duplicate features in a single rule will cause a contradiction. 
Each value of the feature labeling a node is assigned to exactly one edge leaving from the node. 
When a path reaches the depth of $k$ the generator assigns a class to the corresponding leaf.  The class of the leaf is also randomly assigned but in a clever way so as to prevent its parent's feature to be free. The procedure \textit{clever\_random\_select}(C) makes sure that two distinct classes are assigned to the leaves with the same parent.

Once the oracle tree is fully constructed, the generator will output the tree, which can now act as a reference when producing training sets and evaluating decision trees.

\subsection{Training DataSet Generation from Oracles}
After obtaining an oracle from the oracle generator, our next goal is to generate training datasets from the oracles. Our dataset generator follows the rules from the oracle and randomly produces a dataset with a pre-defined number of instances. The generated training datasets  are \emph{fully consistent} with the oracles. 

To better understand possible behaviors of the generator, it is necessary to first investigate the relations between the number of features, the depth, and the number of unique data instances. For $m$ binary-valued features the total number of unique data instances in the training dataset is $2^m$. Observe that each input triggers exactly one rule in a "perfect" tree of depth $k=m$.
If the prescribed depth $k < m$, the features not present in a rule do not impact the classification result. These features are said to be \emph{free}. Note that even though a feature is free in a specific rule, it still cannot be classified as irrelevant. Indeed, different rules contain different feature subsets, and the free features of one rule may well be essential components of other rules. For each rule $\pi$ of length $k<m$ can be triggered by  $2^{m-k}$ unique inputs. 

We consider the generation of  two different types of datasets:  completely random datasets and  uniquely random datasets. A uniquely random dataset has no redundant instance whereas instances can be redundant in completely random datasets. 

\paragraph{Completely Random Dataset generation.} 
Our dataset generation algorithm takes as inputs the number $q$ of (not necessarily distinct) instances to be generated, an oracle $O$, and the feature set $F$. Then it performs $q$ random walks from the root to a leaf of the oracle. Each walk corresponds to a rule and a rule corresponds to a number of some randomly chosen instances (recall that free features in a rule can take any value). The algorithm can walk several times through the same rule, leading to redundant instances in the generated training dataset.

Completely random datasets simulate real-life datasets, in which many inputs are redundant. Depending on the number of instances, it is also possible that some rules have no representation in the dataset. With this type of dataset, we can observe the performance of different decision tree learning algorithms when inferring models with incomplete information.

\paragraph{Uniquely Random Dataset generation.}
Similar to generating completely random datasets, our algorithm for generating uniquely random dataset takes as inputs $q$, $O$ and $F$ with $|F| = m$. Obviously, $q \leq 2^m$. The algorithm could also walk over the rules of the oracle meanwhile ensuring the uniqueness of the produced instances, which is not a simple task. So our algorithm proceeds in two steps. First, a complete dataset $D_{all}$ with all possible instances consistent with the oracle is generated. The first step performs a depth-first  enumeration of rules of the oracle; each time a rule is encountered the corresponding instances are produced as follows: the values of the features in the rule are determined by the label of the edges and free features are iteratively assigned to a distinct  possible value. In the second step, the algorithm randomly chooses  $q$ instances  from $D_{all}$ to produce the uniquely random dataset $D$. The chosen instances are required to be consistent with the maximal number of rules of the oracle to ensure that the maximal number of rules of the oracle is covered by the instances. In other words, let $b$ be the number of rules in oracle $O$. If $q\leq b$ then each instance in $D$ corresponds to a distinct rule in $O$. If $q>b$ then some instances correspond to the same rule in $O$ and differ from each other on the value of the free features in the rule.  

The goal of generating a uniquely random dataset is to provide as much information in the dataset as possible without having redundant inputs. With such a dataset, we are able to discover the effect of having different representations of the same rule on the accuracy of the classification.

\subsection{DOE of Learned Trees Based on Distinguishing Trees}
\label{sec:approach:doe}
Now that we have a training dataset generated based on the oracle and trained the decision trees on the dataset, we determine the DOE to evaluate the degree of correctness of the trained model. 

\begin{algorithm}[!tb]{\textbf{Input:} $O$, $T$
\\
\textbf{Output:} $DOE$}
    \begin{algorithmic}[1]
        \State $ptr_1$ $\leftarrow$ $o_0$, $ptr_2$ $\leftarrow$ $t_0$ \Comment{$ptr_1$ and $ptr_2$ points to root of $O$ and $T$ respectively}
        \State $total$ $\leftarrow$ $0$, $succ$ $\leftarrow$ $0$
        \State \Call{ScanTree}{$ptr_1$, $ptr_2$, $cache$, $total$, $succ$}
        \State $DOE$ $\leftarrow$ $succ / total$
        \State \textbf{return} $DOE$ 
   \\
    \Procedure{ScanTree}{$ptr_1$, $ptr_2$, $cache$, $total$, $succ$}
        \If{$ptr_1$ not leaf and $ptr_2$ not leaf}
            \For{each $vl$ in $V_{(feature(ptr_1))}$}
                \State $ptr_1$ $\leftarrow$ $o_{vl}$ \Comment{move $ptr_1$ to the child node $o_{vl}$}
                \State add(($feature(ptr_1)$, $vl$), $cache$) \Comment{add the feature-value pair to $cache$}
                \If{feature($ptr_2$) in $cache$}
                    \State $vf$ $\leftarrow$ $cache$.get(feature($ptr_2$)) \Comment{get the value of feature($ptr_2$)}
                    \State $ptr_2$ $\leftarrow$ $t_{vf}$ \Comment{move $ptr_2$ to the child node $t_{vf}$}
                \EndIf
                \State \Call{ScanTree}{$ptr_1$, $ptr_2$, $cache$, $total$, $succ$}
            \EndFor
        \ElsIf{$ptr_1$ is leaf and $ptr_2$ not leaf}
            \If{feature($ptr_2$) in $cache$}
                \State $vf$ $\leftarrow$ $cache$.get(feature($ptr_2$))
                \State $ptr_2$ $\leftarrow$ $t_{vf}$
                \State \Call{ScanTree}{$ptr_1$, $ptr_2$, $cache$, $total$, $succ$}
            \ElsIf{feature($ptr_2$) not in $cache$}
                \For{each $vf$ in $V_{(feature(ptr_2))}$}
                    \State $ptr_2$ $\leftarrow$ $t_{vf}$
                    \State \Call{ScanTree}{$ptr_1$, $ptr_2$, $cache$, $total$, $succ$}
                \EndFor
            \EndIf
        \ElsIf{$ptr_1$ not leaf and $ptr_2$ is leaf}
            \For{each $vl$ in $V_{(feature(ptr_1))}$}
                \State $ptr_1$ $\leftarrow$ $o_{vl}$ 
                \State add(($feature(ptr_1)$, $vl$), $cache$)
                \State \Call{ScanTree}{$ptr_1$, $ptr_2$, $cache$, $total$, $succ$}
            \EndFor
        \ElsIf{$ptr_1$ is leaf and $ptr_2$ is leaf}
            \State $total += 1$ \Comment{add this rule to $total$}
            \If{label($ptr_1$) = label($ptr_2$)}
                $succ += 1$ \Comment{add this rule to $succ$}
            \EndIf
        \EndIf
    \EndProcedure
    \end{algorithmic}
    \caption{Equivalence Test}
    \label{algo: equivalence test}
\end{algorithm}

%This recursive process guarantees that no duplication occurs when counting the rules because we are using the oracle as the reference and always trace the inferred model accordingly. Once the process is complete we calculate the degree of equivalence ($DOE$) of the decision tree by dividing the number of consistent rules by the total number of rules.

%%% old text
%%%To demonstrate the evaluation process we consider the following example.
%%% Assume that a dataset $D$ is generated based on the tree in Figure
%%% \ref{fig:perfect tree}, which is the oracle $O$ in our example.  Let a
%%% decision tree learning algorithm produce the model $T$ shown in Figure
%%% \ref{fig:inferred tree} after training on $D$.
%%% suggested text:

Algorithm \ref{algo: equivalence test} performs the equivalence test of a learned tree and an oracle via the computation of the distinguishing tree and the DOE with procedure \textsc{ScanTree}. The procedure maintains two pointers,  with one of the pointers $ptr_1$ tracing the nodes in oracle $O$ and the other pointer $ptr_2$ tracing the nodes in the learned model.  In addition, a cache of values $cache$ stores the features and values visited by the pointer in the oracle tree. The tracing starts at the root node of the distinguishing tree with an empty cache (for stating that features are undefined); then it follows the oracle node by node and rule by rule. While tracing, we record the total number of compatible rules visited $total$, and the number of consistent rules $succ$. No duplication occurs when counting the rules. Finally, it computes the DOE. 

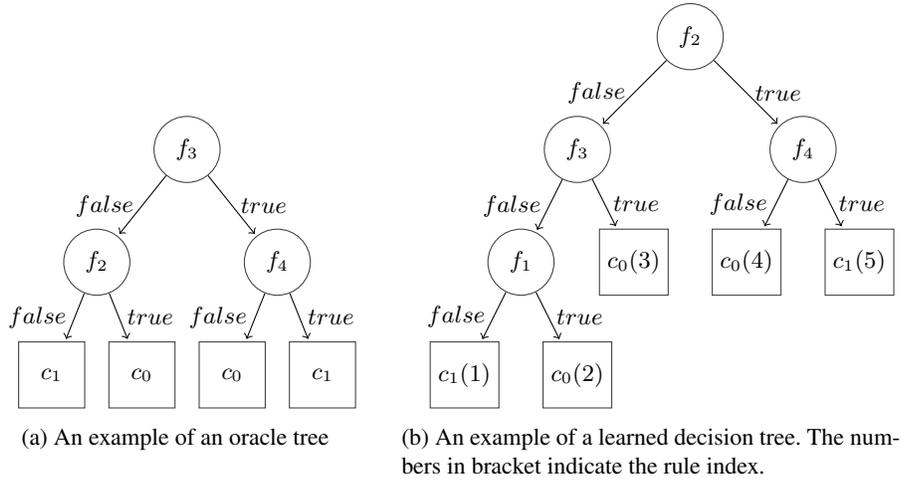
\begin{figure}[t]
 \centering
  \subfloat[An example of an oracle tree]{
     \label{fig:perfect tree}
     \scalebox{1}[1]{
    \begin{tikzpicture}[scale=1.5,shorten >=1pt,->,every node/.style={scale=1}]
    \tikzstyle{vertex}=[circle,draw=black!75,minimum size=25pt]
    \tikzstyle{leaf} = [rectangle,draw=black!75,minimum size=25pt]
 
    \node[vertex] (Q1) at (0,0) 	{$f_3$}  ;
    \node[vertex] (Q2) at (-0.8,-1) 	{$f_2$}  ;
    \node[vertex] (Q3) at (0.8,-1) 	{$f_4$}  ;
    \node[leaf] (Q4) at (-1.2,-2) {$c_1$}  ;
    \node[leaf] (Q5) at (-0.4,-2)	{$c_0$}  ;
    \node[leaf] (Q6) at (0.4,-2) 	{$c_0$}  ;
    \node[leaf] (Q7) at (1.2,-2) 	{$c_1$}  ;
    
    \path (Q1) edge node[left]         {$false$}  	(Q2);
    \path (Q1) edge node[right]         {$true$}  	(Q3);
    \path (Q2) edge node[left]         {$false$}  	(Q4);
    \path (Q2) edge node[right]         {$true$}  	(Q5);
    \path (Q3) edge node[left]         {$false$}  	(Q6);
    \path (Q3) edge node[right]         {$true$}  	(Q7);
     
    \end{tikzpicture}
     }}
     \hfill
  \subfloat[An example of a learned decision tree. The numbers in bracket indicate the rule index.]{
    \label{fig:inferred tree}
    \scalebox{1}[1]{
    \begin{tikzpicture}[scale=1.5,shorten >=1pt,->,every node/.style={scale=1}]
    \tikzstyle{vertex}=[circle,draw=black!75,minimum size=25pt]
    \tikzstyle{leaf} = [rectangle,draw=black!75,minimum size=25pt]
 
    \node[vertex] (Q1) at (0,0) 	{$f_2$}  ;
    \node[vertex] (Q2) at (-1,-1) 	{$f_3$}  ;
    \node[vertex] (Q3) at (1,-1) 	{$f_4$}  ;
    \node[vertex] (Q4) at (-1.5,-2) {$f_1$}  ;
    \node[leaf] (Q5) at (-0.5,-2)	{$c_0(3)$}  ;
    \node[leaf] (Q6) at (0.5,-2) 	{$c_0(4)$}  ;
    \node[leaf] (Q7) at (1.5,-2) 	{$c_1(5)$}  ;
    \node[leaf] (Q8) at (-2,-3) {$c_1(1)$}  ;
    \node[leaf] (Q9) at (-1,-3)	{$c_0(2)$}  ;
    
    \path (Q1) edge node[left]         {$false$}  	(Q2);
    \path (Q1) edge node[right]         {$true$}  	(Q3);
    \path (Q2) edge node[left]         {$false$}  	(Q4);
    \path (Q2) edge node[right]         {$true$}  	(Q5);
    \path (Q3) edge node[left]         {$false$}  	(Q6);
    \path (Q3) edge node[right]         {$true$}  	(Q7);
    \path (Q4) edge node[left]         {$false$}  	(Q8);
    \path (Q4) edge node[right]         {$true$}  	(Q9);

    \end{tikzpicture}
    }}
  \caption{Examples of an oracle tree and a learned tree}
  \label{fig:ex2}
\end{figure}

\begin{table}[]
    \centering
    \begin{tabular}{|c|c|c|}
        \hline
         Rules & Compatible Rules & Consistent Rules \\ \hline
         $f_3 = false$ and $f_2 = false$ & rule $1$, rule $2$ & rule $1$ \\ \hline
         $f_3 = false$ and $f_2 = true$ & rule $4$, rule $5$ & rule $5$ \\ \hline
         $f_3 = true$ and $f_2 = false$ & rule $3$, rule $4$ & rule $3$, rule $4$ \\ \hline
         $f_3 = true$ and $f_2 = true$ & rule $3$, rule $5$ &  rule $5$\\ \hline
    \end{tabular}
    \caption{Rules in the learned tree that are compatible and consistent with the oracle rules}
    \label{tab:equivtest}
\end{table}

Let us provide an example of the execution of Algorithm~\ref{algo: equivalence test} with the oracle  in Figure \ref{fig:perfect tree} and the tree in Figure \ref{fig:inferred tree} obtained by applying a learning algorithm on a dataset generated from the oracle. We refer to a triplet $\langle prt_1, ptr_2, cache \rangle$ as the \textit{scanner} data structure. The initial scanner is $\langle f_3, f_2, \{\} \rangle$ where  $ptr_1=f_3$ corresponds to the root of the oracle, $ptr_2 =f_2$ corresponds to the root of the learned tree, and $cache$ is empty.  $total$ and $succ$ are both initialized with $0$.
The algorithm first proceeds on the leftmost rule of the oracle  by choosing the value $false$ for $f_3$ and moving the pointer to the child with $f_2$ as the label. We update $scanner$ with the new feature and the selected value of the visited feature, which gives $\langle f_2, f_2, \{f_3 = false\}\rangle$. Because the feature label $f_2$ of the node that $ptr_2$ points to does not exist in $cache$ yet, $ptr_2$ stays at the same position. We then move the pointer $ptr_1$ to the leaf of this rule, which is $c_1$, by choosing the value $false$ for $f_2$. The updated $scanner$ is $\langle c_1, f_2, \{f_3 = false, f_2 = false\}\rangle$. We notice that $f_2$ is present in $cache$ now, so $ptr_2$ should move to the child satisfying $f_2 = false$, which is the node labeled with $f_3$. Again since $f_3$ exists in $cache$, $ptr_2$ moves to the node labeled $f_1$. The new $scanner$ is updated as $\langle c_1, f_1, \{f_3 = false, f_2 = false\}\rangle$.  Now, because $f_1$ is not present in the cache, $ptr_2$ has to visit every child of the current node. When $f_1 = false$, $ptr_2$ reaches the leaf and the class label is $c_1$. Since both pointers reach the leaves on their respective tree and the resulting class labels are the same, we add this rule to both $total$ and $succ$. We say rule with index $1$ on the learned tree (see Figure~\ref{fig:inferred tree})is consistent with the leftmost rule in the oracle. Then we move $prt_2$ to the next child of $f_1$ where $f_1 = true$. Again, both pointers reached the leaf node so we shall add this rule to $total$. However, because the leaves have different class labels, we do not increment $succ$. Hence, for the leftmost rule in the oracle tree, two rules in the learned tree are compatible with it, but only one of them is consistent with the oracle. We repeat this process rule-by-rule and so we are able to scan both trees thoroughly. Table \ref{tab:equivtest} shows the rules in the learned tree that are compatible and consistent with each oracle rule. By summing the number of compatible rules, we get $total = 8$. Similarly, we calculate $succ$ by summing up the number of consistent rules, which is $5$. Therefore, the overall DOE for this learned tree w.r.t the oracle is $0.625 = 5/8$.

\section{Empirical Assessment of the Proposed Approach}\label{sec:exp}
We empirically assess the proposed approach to evaluating learning algorithms for decision trees. 
To this end, we conduct experiments with decision tree learning tools  and a prototype tool we have developed. The decision tree learning algorithms include four heuristic-based algorithms namely
ID3 \cite{quinlan1986induction}, J48 (a WEKA implementation of C4.5 \cite{Quinlan1993}), simpleCART (a WEKA implementation of CART \cite{Breiman1984}), and RandomTree~\cite{frank2009weka}, and an exact algorithm which infers optimal decision trees namely InferDT~\cite{DBLP:conf/aaai/Avellaneda20}.
%two decision tree learning algorithms we have implemented. \omer{Add short a description of the two algorithms} 
The prototype tool includes three main functionalities: oracle generation, data generation from oracles, and computation of the accuracy of learned trees based on the comparison between them and the oracles.

ID3, which stands for Iterative  Dichotomiser  3, is a basic yet powerful decision tree learning algorithm designed in 1986 by Ross Quinlan \cite{quinlan1986induction}. The idea of ID3 is to construct a decision tree by using a heuristic-based greedy search algorithm to test each feature in the data subset at each node. The heuristic function, called \emph{Information Gain}, minimizes the information entropy of different classes. C4.5 \cite{Quinlan1993} is developed based on ID3 so it has a similar design which also employs the concept of information entropy when constructing the decision tree. But because information gain tends to favor features with a larger value set, C4.5 uses the \emph{Information Gain Ratio} \cite{quinlan2014c4} to mitigate this problem. CART \cite{Breiman1984} was first introduced in 1984 and it has been widely used ever since. Differing from ID3 and C4.5, CART embraces another definition of impurity namely \emph{Gini impurity}. Standard C4.5 and CART include a \emph{pruning} process to reduce the effect of data uncertainty. However, in our approach, the datasets generated by the oracle are deterministic and contain no noise, so we are evaluating the performance of these learning algorithms without the pruning stage. 

In recent years, exact model inferences are getting increased attention. Algorithms like InferDT \cite{DBLP:conf/aaai/Avellaneda20} are aiming to find the optimal decision trees consistent with the learning datasets. They are known for their ability to produce very accurate models, but very few research studies have been done comparing them to the heuristic-based decision tree algorithms. In our experiments, we evaluate the quality of these optimal decision trees by considering only the maximum depth as a criterion of optimality. 

%%%%% COPY / PASTE FROM THESIS PAPER %%%%%%%%%%%%%%%

We are aiming to answer the following questions during our empirical experiments:

\begin{description}
\item[Question 1] With the same number of features and depth in the oracle, what is the relation between the number of data instances in the training set and the degree of correctness of the model inferred by the learning algorithms?

\item[Question 2] With the same number of features, how does the  depth of the oracles impact the performance of decision tree learning algorithms in terms of DOE?

\item[Question 3] With the same number of features and depth in the oracle and the same number of instances in the training set, which decision tree learning algorithm infers the most accurate model?

\item[Question 4] With the same number of features and depth in the oracle, what is the difference when training on completely random datasets and on uniquely random datasets?
\end{description}

To answer the above questions we perform a number of atomic experiments. Each such an experiment takes three parameters as inputs: the number $n$ of features, the maximal depth $k$ of an oracle, with $1\leq k\leq n$, and the size $m$ of the training dataset. An experiment proceeds in four steps. First,  we generate an oracle with Algorithm~\ref{algo: rule generator}. A generated oracle uses $n$ features and the length of each of its rules is $k$. Two rules in the same oracle can use different features and the order of occurrence of the features on two distinct rules can be different. In the second step, we randomly generate a training dataset with $m$ inputs from the oracle. Third, for each generated training dataset we use the $L$-th learning algorithm under evaluation to generate the $L$-th learned tree. In the fourth step, we determine the DOE of the generated learned trees by performing an equivalence test for each learned tree against the oracle.

For each set of input values ($n$, $k$, $m$, $L$), we perform a number $tl$ of experiments and average the DOE values obtained from the equivalent tests. The purpose of calculating average DOE is to minimize the potential performance bias when the respective algorithm trains on a specific dataset. In our experiment, we set $tl$ to $100$ when evaluating the performance of ID3, J48, simpleCART, and RandomTree; on the other hand, we set $tl$ to $20$ when inferring tree models using InferDT because the results from optimal tree inference are more stable.

\begin{figure*}[!t]
    \centering
    \begin{subfigure}[t]{0.49\textwidth}
        \centering
        \includegraphics[width=\linewidth]{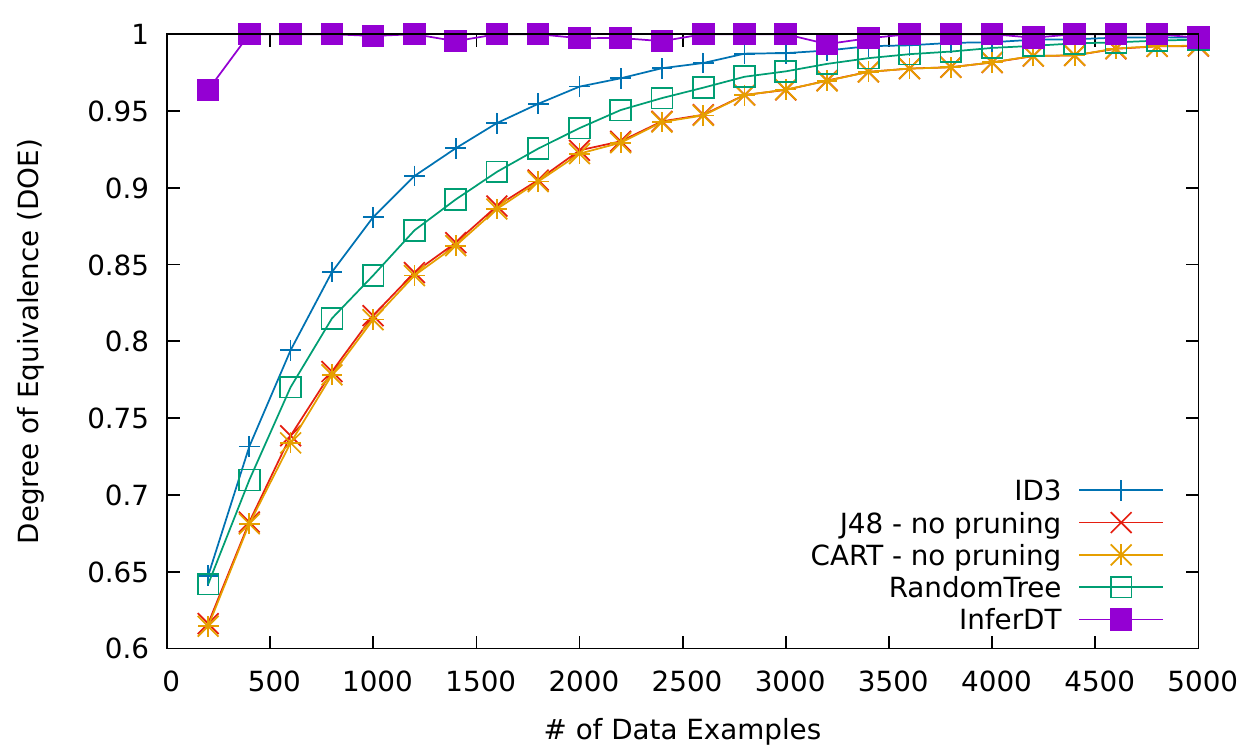}
        \caption{Depth = 5}
    \end{subfigure}%
    ~ 
    \begin{subfigure}[t]{0.49\textwidth}
        \centering
        \includegraphics[width=\linewidth]{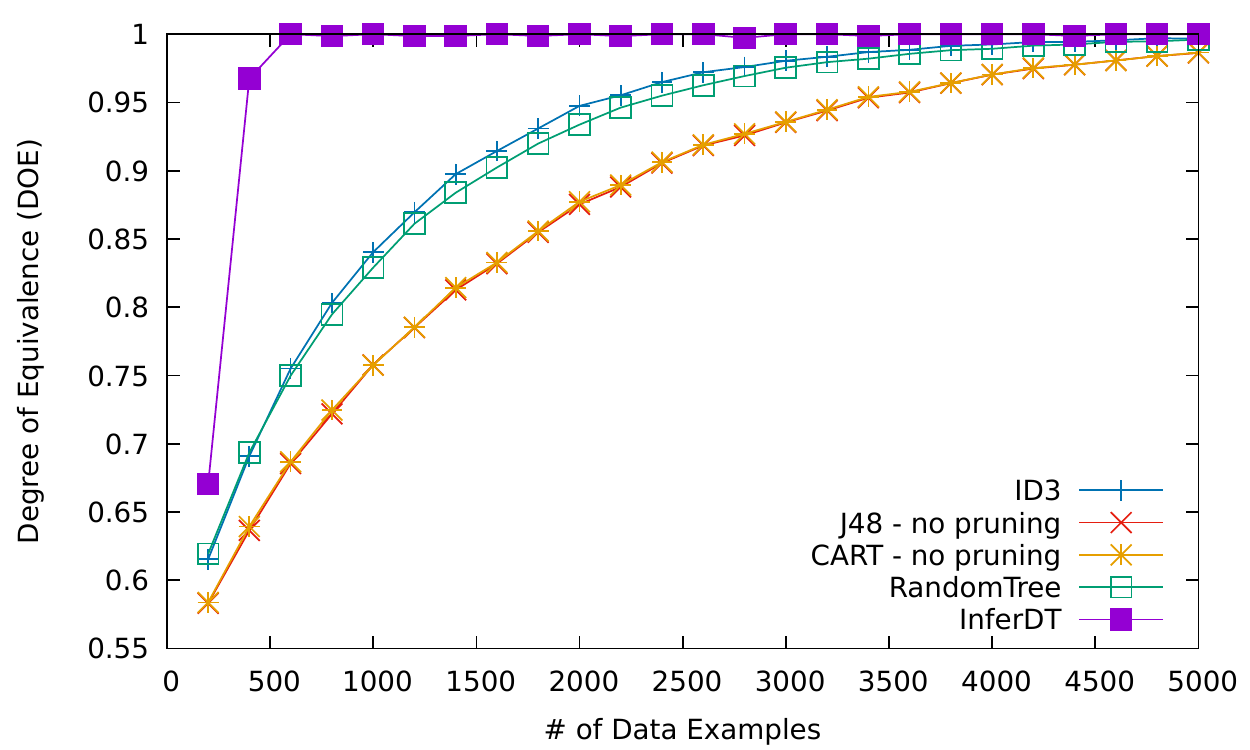}
        \caption{Depth = 6}
    \end{subfigure}
    \newline
    \begin{subfigure}[t]{0.49\textwidth}
        \centering
        \includegraphics[width=\linewidth]{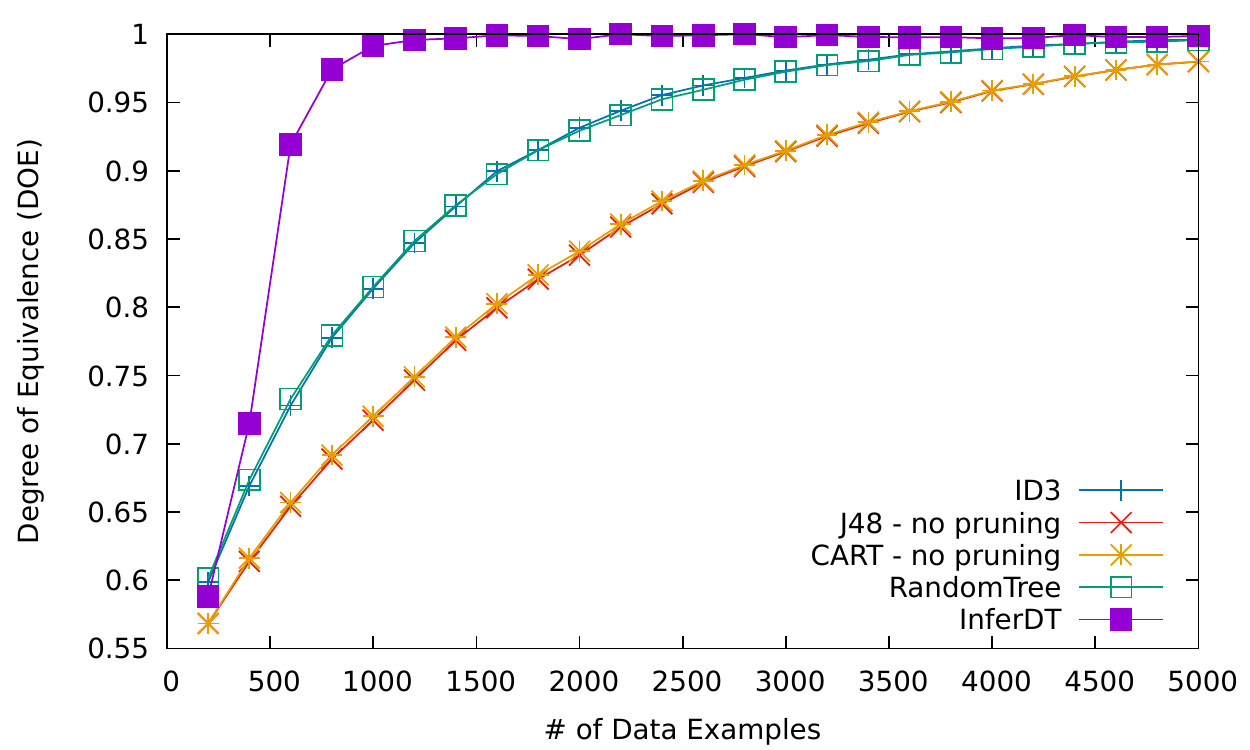}
        \caption{Depth = 7}
    \end{subfigure}%
    ~ 
    \begin{subfigure}[t]{0.49\textwidth}
        \centering
        \includegraphics[width=\linewidth]{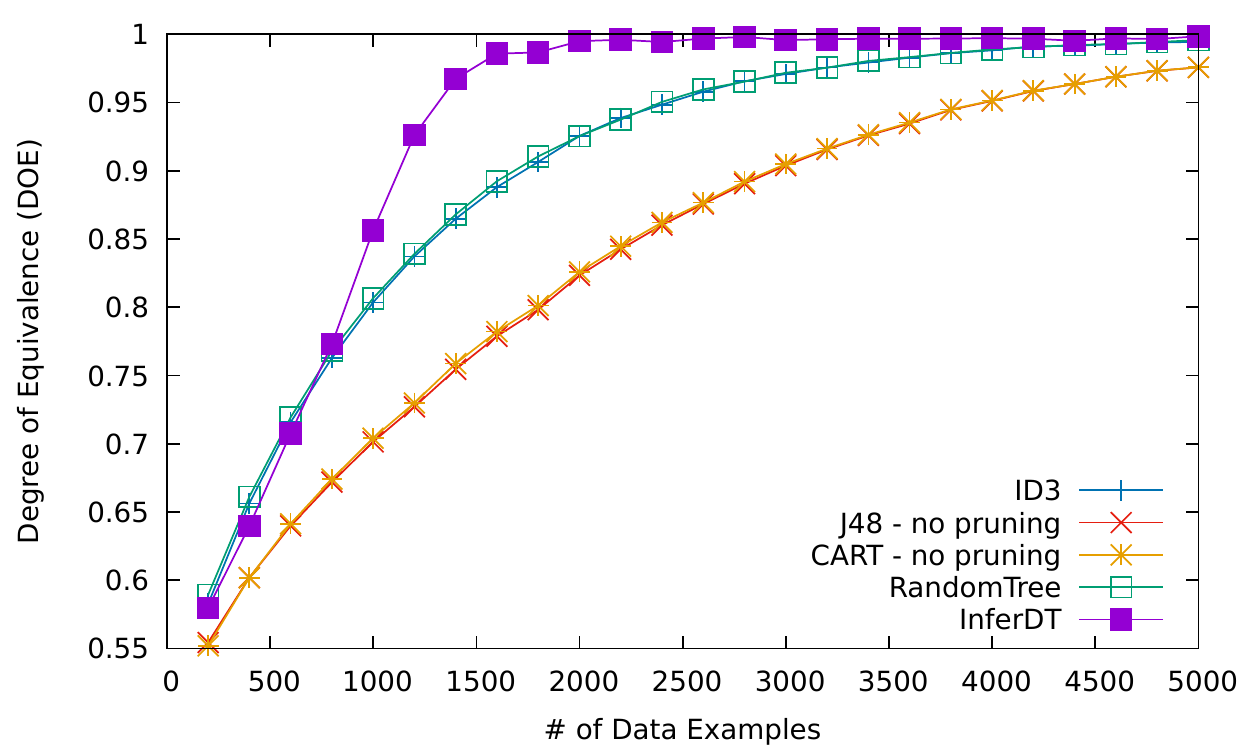}
        \caption{Depth = 8}
    \end{subfigure}
    %\newline
    %\begin{subfigure}[t]{0.5\textwidth}
    %    \centering
    %    \includegraphics[height=1in]{figure-label.png}
    %\end{subfigure}
    \caption{DOE comparison for decision tree learning algorithms trained on completely random datasets with 10 features and binary values}
    \label{fig:DOE completely random}
\end{figure*}

Figure~\ref{fig:DOE completely random} uses line plots to illustrate the performance difference between decision tree learning algorithms for different input depth, where the number of features $n$ is fixed to $10$, and the training is done on completely random datasets. In these plots, each point $(x, y)$ represents the average DOE values $y$ obtained from the equivalence tests between the oracle and the tree constructed by the learning algorithm trained on a dataset with the given size $x$. The depth input ranges from $5$ to $8$, aiming to observe the performance difference of each decision tree learning algorithm when the oracle tree grows deeper. 

As shown in the plots, when the number of features and depth is fixed the DOE score increases as the size of the training set gets larger (\textbf{Question 1}). For the heuristic-based decision trees, the line plots show a logarithmic-like increase, whereas the DOE scores of InferDT increase near-linearly with a very steep slope and exceed $99\%$ when the number of inputs in the datasets is relatively small. 

We also observe that, with the same number of features and as the depth of the oracle increases, the learning algorithms need larger training sets to train on in order to achieve the same DOE score as before (\textbf{Question 2}). To take ID3 as an example, it requires $1200$ random inputs to infer a model with a $90\%$ DOE score when depth is $5$. However, when the depth is set to $8$, a dataset with $1800$ inputs is needed to train a model with the same DOE score. J48 and simpleCART are more sensitive to depth increase. When the oracle deepens from $5$ to $8$, J48 and simpleCART need training sets with nearly double the size (from $1800$ to $3000$) to produce a model with $90\%$ DOE score. It is also noticeable that the curves are flattened when the depth escalates. 

Based on the graphs, InferDT clearly outperforms all the heuristic-based learning algorithms because it requires fewer inputs to infer an accurate model (\textbf{Question 3}). However, because the computational time increases exponentially as the oracle tree gets deeper, this algorithm takes much longer to produce a tree model. ID3 shows significantly better results when compared to J48 and simpleCART despite it being the earliest member of the decision tree algorithm family. It also performs better than RandomTree when the depth of oracle is small; however, because ID3 is more sensitive to the depth increase, the performance difference between the two algorithms becomes very small when the depth grows. The performance of J48 and simpleCART are very similar since we observe that their curves overlap each other in every plot.

\begin{figure*}[t!]
    \centering
    \begin{subfigure}[t]{0.49\textwidth}
        \centering
        \includegraphics[width=\linewidth]{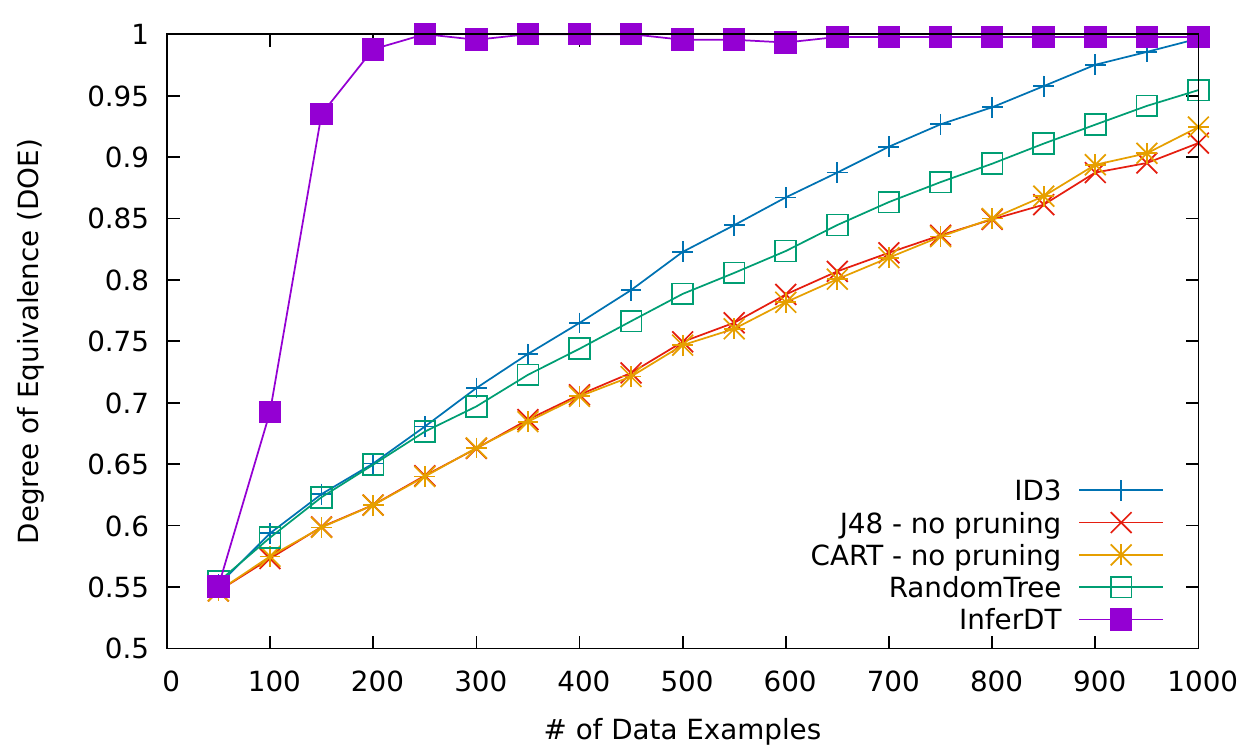}
        \caption{Depth = 5}
    \end{subfigure}%
    ~ 
    \begin{subfigure}[t]{0.49\textwidth}
        \centering
        \includegraphics[width=\linewidth]{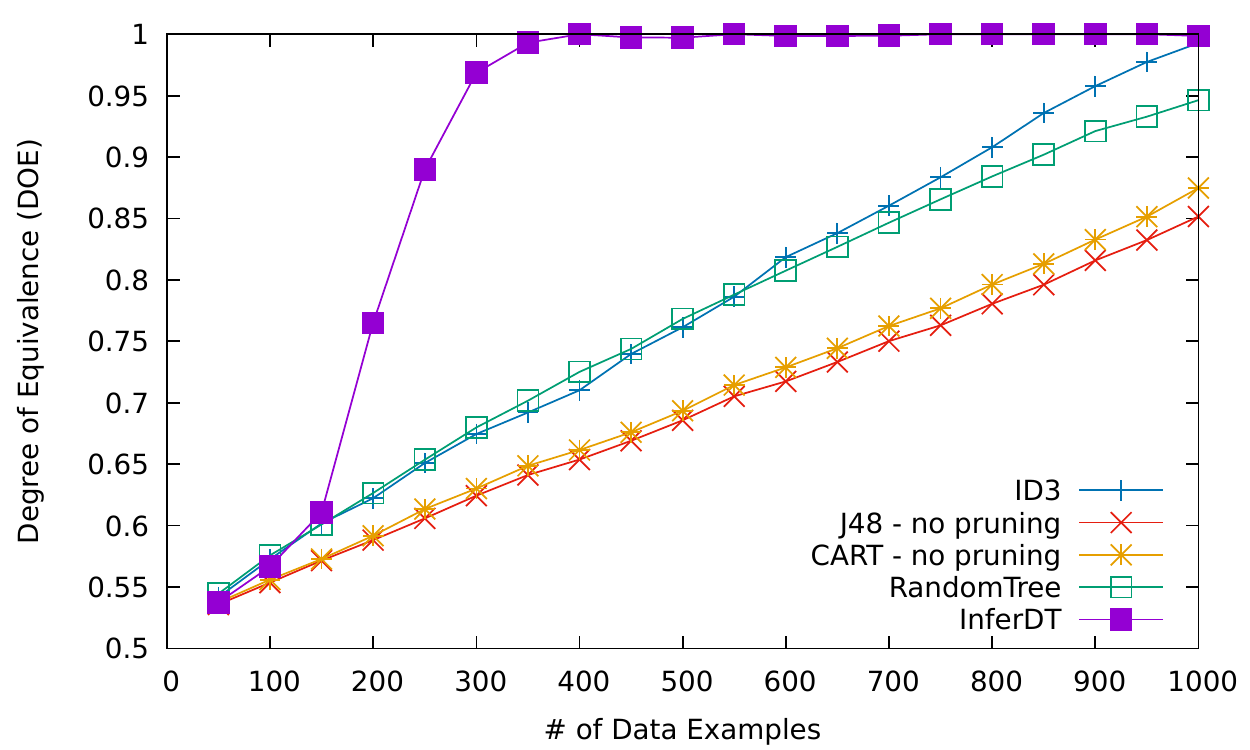}
        \caption{Depth = 6}
    \end{subfigure}
    \newline
    \begin{subfigure}[t]{0.49\textwidth}
        \centering
        \includegraphics[width=\linewidth]{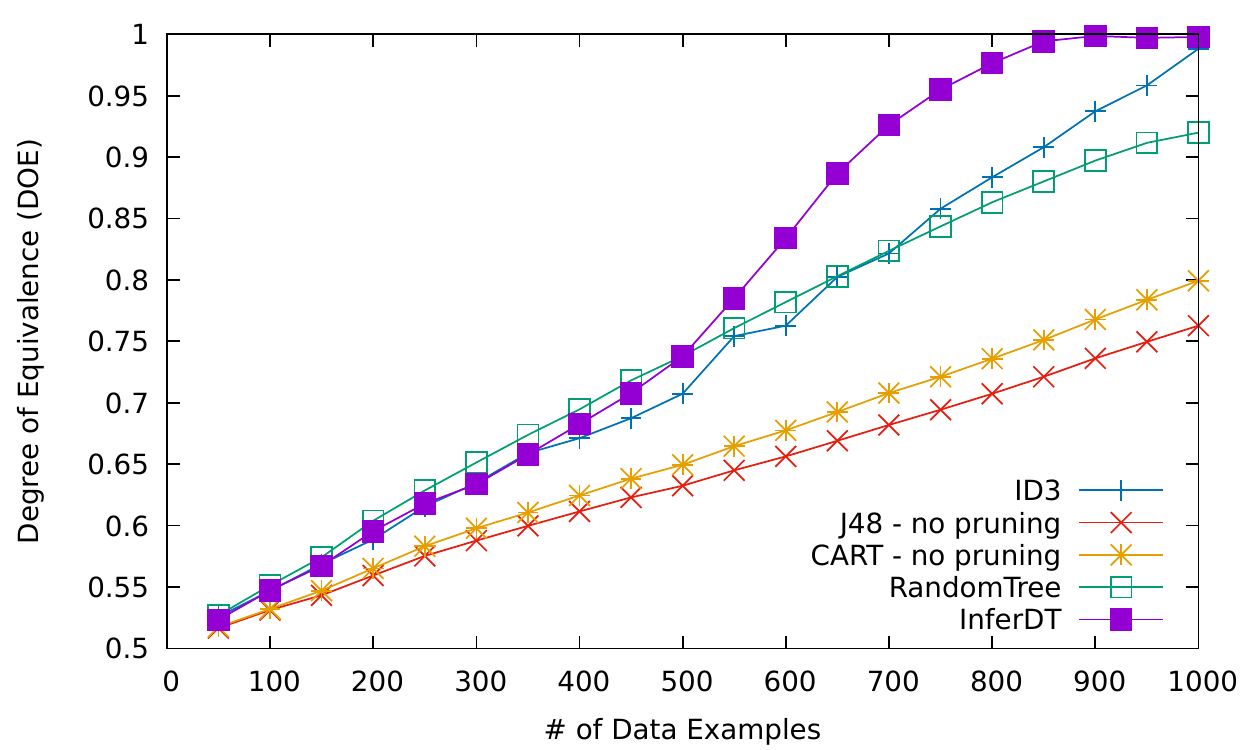}
        \caption{Depth = 7}
    \end{subfigure}%
    ~ 
    \begin{subfigure}[t]{0.49\textwidth}
        \centering
        \includegraphics[width=\linewidth]{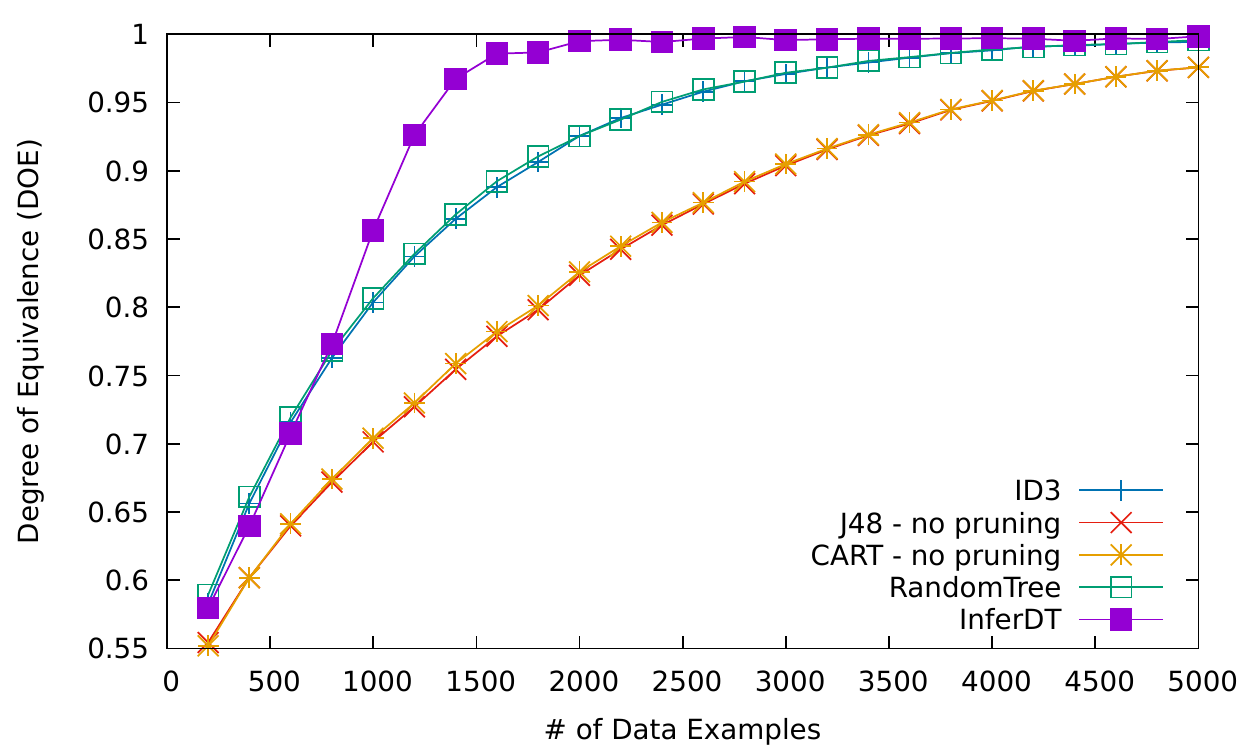}
        \caption{Depth = 8}
    \end{subfigure}
    %\newline
    %\begin{subfigure}[t]{0.5\textwidth}
    %    \centering
    %    \includegraphics[height=1in]{figure-label.png}
    %    %\caption{Algorithms}
    %\end{subfigure}
    \caption{DOE comparison for decision tree learning algorithms trained on uniquely random datasets with 10 features and binary values}
    \label{fig:DOE uniquely random}
\end{figure*}

To answer \textbf{Question 4}, we also performed a series of experiments using uniquely random datasets. Similar to the above-mentioned experiments, we evaluate the objective decision tree algorithms by training them on the generated training sets. We then compare the DOE score computed by the equivalence tests and plot line graphs to illustrate the performance difference for inputs of various depths. Figure~\ref{fig:DOE uniquely random} shows the results of these experiments. The number of features is again set to $10$ for comparable results. 

One major difference between the results obtained from these two sets of experiments is the size of the dataset. For $10$ features with binary values, the total number of unique inputs is $1024$ ($2^{10}$). Hence, it is impossible for the number of inputs in the uniquely random datasets to be more than $1024$. On the other hand, completely random datasets contain redundant inputs, which means a larger dataset is required to represent the same amount of information as in the uniquely random dataset. 

Another observation from comparing the two sets of results is the difference in the shape of the curves in the line plots. In contrast to the logarithmic-like curves, when the heuristic-based decision tree learning algorithms train on uniquely random datasets, the line plots are approximately linear. The curve of InferDT shapes distinctively. When the datasets are small, the curves are roughly straight and the slopes are similar to the slopes of ID3 and RandomTree curves; however, when the size of the datasets passes a critical number (i.e., $150$ when the depth is set to $6$, $300$ when the depth is $7$, or $500$ when the depth is $8$), the increase of the DOE values accelerates and the shape of the curves becomes logarithmic-like. 

It may seem odd that InferDT does not have advantages over heuristic-based algorithms when training on datasets with a small number of examples. Yet, the reason is rather simple: small datasets do not have enough inputs to fully represent the entire oracle model. For an oracle with a depth of $k$, at least $2^k$ inputs are needed in the datasets to represent every rule in the oracle. As the depth $k$ gets larger, the minimum number of examples for full oracle representation grows exponentially. Without enough data providing information about the oracle, the exact model inferred by InferDT would not be equivalent to the oracle model. Note that when the size of the datasets grows over a critical number, the performance of InferDT improves drastically.

Overall, InferDT shows the best performance among the learning algorithms under test (\textbf{Question3}). Even if InferDT produces similar DOE values as ID3 and RandomTree when the training sets contain a limited number of instances, it quickly surpasses the other learning algorithms as the size of the datasets grows. ID3 is better than the other heuristic-based learning algorithms, especially when training on large datasets. RandomTree also infers very accurate models in general. It even defeats ID3 when dealing with small datasets generated by deep oracle models. J48 and simpleCART again produce similar results, but simpleCART performs slightly better when the oracle tree is deeper. 
%%%%%%%%%%%%% END COPY / PASTE FROM THESIS PAPER %%%%%%%%%%%%%%%%%%

\section{Conclusion}\label{sec:concl}
We proposed a novel approach to evaluating decision tree learning algorithms. The approach is based on the empirical comparison of oracles trees with learned decision trees. The decision trees are learned from datasets randomly generated from oracle trees. The evaluation result is independent of a specific dataset. We evaluated five decision tree learning algorithms, namely ID3, J48, simpleCART, RandomTree, and InferDT. The preliminary evaluation results show that, when training on deterministic datasets with no noise, InferDT produces the most accurate model. In the family of heuristic-based decision trees, ID3 and RandomTree have the best performance, where ID3 performs slightly better than RandomTree. The results also show the effectiveness of the proposed evaluation method. By using DOE as the metric, it successfully distinguished the performance difference between learning algorithms. 

For future research, we plan on enhancing our framework to consider noisy data, which involves generating non-deterministic datasets. We are expecting J48 and simpleCART to show better performance in this context because of their pruning process. We also intend to apply this approach to evaluate feature selection techniques. Indeed, when the depth of an oracle is smaller than the number of available features, each rule would have "free features" that do not contribute to assigning class labels. These free features are \emph{irrelevant} to this rule. Based on these free features we are expecting then to be able to rank features based on their relevancy and correctly identify the ones that are irrelevant overall. We would also like to examine the relation between the size of the dataset and the performance of these feature selection techniques in terms of their ability to recognize trivial features. We also plan to investigate more complex dataset generation approach, which can be inpire by the work in ~\cite{AichernigB0HPRR19}.

%\bibliographystyle{plain}
%\bibliography{bibi}

\end{document}